\documentclass{article}

\usepackage[final]{ewrl_2026}

\usepackage[utf8]{inputenc}
\usepackage[T1]{fontenc}
\usepackage[hypertexnames=false]{hyperref}
\usepackage{url}
\usepackage{booktabs}
\usepackage{multirow}
\usepackage{amsmath}
\usepackage{amsthm}
\usepackage{amsfonts}
\usepackage{nicefrac}
\usepackage{microtype}
\usepackage{xcolor}
\usepackage{graphicx}
\usepackage{caption}
\usepackage{subcaption}
\usepackage{float}
\usepackage{wrapfig}

\title{EfficientTDMPC: Improved MPC Objectives
for Sample-Efficient Continuous Control}

\author{%
  Thomas Evers \\
  TU Delft \\
  \texttt{thomasevers9@gmail.com} \\
  \And
  Cristian Meo \\
  LatentWorlds AI \\
  \And
  Wendelin Böhmer \\
  TU Delft \\
  \AND
  Justin Dauwels\thanks{Shared last authors.} \\
  TU Delft \\
  \And
  Yaniv Oren$^*$ \\
  TU Delft
}

\begin{document}

\maketitle


\begin{abstract}
We introduce EfficientTDMPC, a sample-efficient model-based reinforcement learning method for continuous control built on the TD-MPC family of algorithms. Central to this family is a planner that aims to find an action sequence that maximizes the estimated return. The return is estimated using a learned model and value networks, each of which can introduce error. EfficientTDMPC introduces three contributions that improve performance by aiming to reduce this error.
First, we introduce a multi-horizon planning objective that evaluates the value at different rollout depths and averages them. Second, to our knowledge we are the first to train a value-equivalent dynamics ensemble. Our improved objective then averages over rollouts from multiple dynamics heads.
Third, we add pessimistic reanalyze for tasks that can terminate early.
Applying our contributions to a recent baseline (BMPC) yields EfficientTDMPC, which to our knowledge is the new state of the art in sample efficiency on HumanoidBench and the DeepMind Control Suite, reaching BMPC's final aggregated performance using 57\% fewer environment steps.

\end{abstract}


\vspace{-1em}
\section{Introduction}

\noindent
\begin{minipage}[t]{0.4\textwidth}
\vspace{-0em}
Model-based reinforcement learning (MBRL) methods that learn a world model and use it for planning have achieved strong sample efficiency in continuous control \citep{yu2020mopo,janner2019mbpo,chua2018pets,hansen2022tdmpc,hansen2024tdmpc2}, with applications spanning robotics, autonomous vehicles, and spatial understanding and reasoning~\citep{hafner2020dreamer, hafner2025dreamerv3, meo2025maskedgenerativepriorsimprove}. 
The TD-MPC family \citep{hansen2022tdmpc} has been very successful in this domain by introducing a latent dynamics model with an MPC planner to find action sequences that maximize a bootstrapped return estimate.
\end{minipage}%
\hfill
\begin{minipage}[t]{0.5\textwidth}
\vspace{-3em}
\centering
\includegraphics{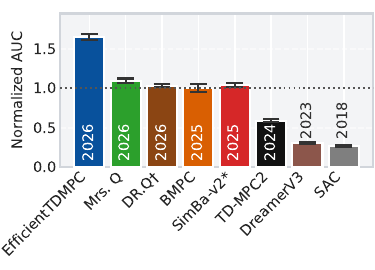}
\captionof{figure}{Normalized mean area under the learning curve (AUC) on HumanoidBench (14 tasks) and the DeepMind Control Suite (28 tasks) over the first half of their full step budget. Bars are relative to BMPC. EfficientTDMPC outperforms the state of the art by over 45\%.
 $^*$$^\dagger$Further details in Appendix~\ref{app:generating-performance-figures}.}
\label{fig:intro_auc}
\end{minipage}

Several recent works in the family, like TD-M(PC)$^2$ \citep{lin2025tdmpc-squared}, BOOM \citep{zhan2025boom} and BMPC \citep{wang2025bmpc}, then distill this planner into the policy network, relying even more strongly on planning.
The planner's purpose is to find action sequences that maximize the estimated return. This estimate is created by rolling out the learned dynamics and reward models and then bootstrapping with a value network. However, when the learned model or value network is inaccurate, the planner can exploit these inaccuracies to find action sequences that are estimated to yield a higher return than they actually do in the environment \citep{chua2018pets, janner2019mbpo}. 

EfficientTDMPC introduces several contributions that improve performance by aiming to reduce the error in this return estimate.
Firstly, we introduce a multi-horizon planning objective: we average the return estimates across different rollout depths, which has been successful in MCTS \citep{browne2012mcts, silver2017alphazero, schrittwieser2020muzero, oren2025tsmcts} but is novel in MPC to our knowledge. 
Secondly, to the best of our knowledge, we are the first to train a value-equivalent dynamics ensemble. We then use that ensemble during planning by  averaging the return across all dynamics models which aims to prevent the exploitation of inaccurate models \citep{chua2018pets, janner2019mbpo, yu2020mopo}. 
Thirdly, we introduce pessimistic reanalyze, which penalizes action sequences for which the return estimate is uncertain when creating policy targets. 
We also add practical improvements such as cheaper planning during reanalyze, which we find reduces wall-clock training time without sacrificing performance. Overall, to the best of our knowledge, EfficientTDMPC is the new state of the art in sample efficiency on HumanoidBench and the DeepMind Control Suite. It also reaches the aggregate baseline (BMPC) performance in 57 \% fewer environment steps without significantly increasing wall-clock training time.
\begin{itemize}
    \item \textbf{Improved planner objectives.} We introduce three changes to the planner objective and show that they contribute significantly to performance: a multi-horizon planning objective, a value-equivalent dynamics ensemble, and pessimistic reanalyze.
    \item \textbf{Pipeline improvements.} Per-step replay insertion, horizon increase and a reduced reanalyze budget.
    \item \textbf{Performance.} To the best of our knowledge, EfficientTDMPC is the new state of the art in sample efficiency on HumanoidBench and the DeepMind Control Suite.
\end{itemize}


\section{Background}
\label{sec:background}

\textbf{Problem setting and notation:}
We consider continuous-control problems modeled as discounted Markov decision processes $(\mathcal S,\mathcal A,P,r,\gamma)$. Here, \(\mathcal{S} \subseteq \mathbb{R}^n\) and \(\mathcal{A} \subseteq \mathbb{R}^m\) denote the state and action spaces, where \(s \in \mathcal{S}\) are states and \(a \in \mathcal{A}\) are actions, \(P : \mathcal{S} \times \mathcal{A} \to \Delta(\mathcal{S})\) is the state-transition kernel, \(r : \mathcal{S} \times \mathcal{A} \to \mathbb{R}\) is the reward function, and \(\gamma \in [0,1)\) is the discount factor. A policy \(\pi : \mathcal{S} \to \Delta(\mathcal{A})\) maps states to action distributions. We define the value of a policy at state \(s\) as the expected discounted return when starting from \(s\) and following \(\pi\) thereafter,
\begin{equation}
\begin{aligned}
V^\pi(s) = \mathbb{E}_{\pi, P} \Big[
&\sum_{t=0}^{\infty} \gamma^t r(s_t, a_t)
\Bigm| s_0 = s,\; a_t \sim \pi(\cdot \mid s_t), \\
&\qquad s_{t+1} \sim P(\cdot \mid s_t, a_t) \Big].
\end{aligned}
\end{equation}

\textbf{TD-MPC family:}
Our method builds on the TD-MPC family of works \citep{hansen2022tdmpc,hansen2024tdmpc2} which use a strong planner to find an action sequence that maximizes the estimated return. The estimated return of an action sequence is then obtained by rolling out a latent world model and bootstrapping with a value network. The planner is used both for acting in the environment and for distilling its behavior into a policy network. Several follow-up works like TD-M(PC)$^2$, BOOM, and BMPC then distill this planner into the policy, relying even more strongly on the planner.
We build on BMPC \citep{wang2025bmpc} specifically, since it is a recent contribution that has shown strong empirical performance.

\textbf{Latent world model:}
Table~\ref{tab:components} lists the model components of BMPC.

\begin{table}[H]
\centering
\caption{Model components of BMPC}
\label{tab:components}
\small
\begin{tabular}{@{}lll@{}}
\toprule
\textbf{Component} & \textbf{Notation} & \textbf{Role} \\
\midrule
Encoder           & $h_{\theta}$                   & Maps observation $s_t$ to latent $z_t = h_{\theta}(s_t)$ \\
Dynamics model    & $f_{\theta}$                   & Predicts next latent: $z_{t+1} = f_{\theta}(z_t, a_t)$ \\
Reward model      & $R_{\theta}$                   & Predicts scalar reward from a latent-action pair \\
Policy (prior)    & $\pi_{\theta}(\cdot \mid z)$      & Stochastic policy used to guide planning \\
Value network     & $V_{\theta,i}(z)$                 & Ensemble of state-value networks $i \in \{1,.., N_v\}$ \\
\bottomrule
\end{tabular}
\end{table}

Given an observed state $s_t$, the encoder produces the latent representation $z_t = h_{\theta}(s_t)$.  The dynamics head is then trained to predict the next latent $\hat{z}_{t+1} = f_{\theta}(z_t, a_t)$ given an action $a_t$. The reward and value networks can be evaluated on encoded or rolled-out latents; as detailed below, they are trained primarily on dynamics rollouts. We use $X_{\theta,i}$ to denote the $i$-th ensemble member, and use the notation $X_\theta$ for the average over the ensemble.

\textbf{Planner objective}
\label{sec:q-estimates}
In the TD-MPC family, a planner optimizes an action sequence of length $H$ to maximize the expected return. The exact objective used is the state-action value of the action sequence, which is defined as the expected return of executing the action sequence $\mathbf{a}_{t:t+H-1}$ starting from $s_t$, and then following policy $\pi$ thereafter.
\begin{equation}
\begin{aligned}
Q_H^\pi(s_t, \mathbf{a}_{t:t+H-1})
&= \mathbb{E}_{P} \Big[
\sum_{u=0}^{H-1} \gamma^u r(s_{t+u}, a_{t+u})
+ \gamma^H V^\pi(s_{t+H}) \\
&\qquad \Big|\; s_t,\; \mathbf{a}_{t:t+H-1},\;
s_{t+u+1} \sim P(\cdot \mid s_{t+u}, a_{t+u})
\Big].
\end{aligned}
\end{equation}

Since direct access to this quantity is not available, the TD-MPC family estimates the state-action value by rolling out its learned dynamics and reward models, and bootstrapping with the value network at the end of the rollout. Specifically, given a latent state $z_t$ and an action sequence $\mathbf{a}_{t:t+H-1}$, the state-action value estimate is defined as
\begin{equation}
\begin{aligned}
\hat{Q}^\pi_H(z_t, \mathbf{a}_{t:t+H-1})
&=
\sum_{u=0}^{H-1}\gamma^u R_{\theta}(\hat{z}_{t+u}, a_{t+u})
+
\gamma^H V^\pi_{\theta}(\hat{z}_{t+H}), \\
\hat{z}_{t+u+1} &= f_{\theta}(\hat{z}_{t+u}, a_{t+u}),
\end{aligned}
\end{equation}
where $\hat{z}_t = z_t$.
This state-action value estimate, which we also call the return estimate, is then used as the objective for the planner. 

Specifically, MPPI is used to solve for the optimal action sequence:
\begin{equation}
\mathbf{a}_{t:t+H-1}^{*}
=
\arg\max_{\mathbf{a}_{t:t+H-1}}
\hat{Q}^\pi_H(z_t, \mathbf{a}_{t:t+H-1})
\end{equation}

Details left out about MPPI and this process can be found in the TD-MPC2 paper \citep{hansen2024tdmpc2}. Here the first step of the optimal action sequence $\mathbf{a}_{t}^{*}$ can be considered as the action sampled from the planning policy $\pi_{\mathrm{plan}}$. All members of the TD-MPC family sample from this expert policy $\pi_{\mathrm{plan}}$ to act in the environment.

\textbf{Value-equivalent training}
The TD-MPC family is part of the value-equivalent (MuZero-style) family which trains the reward and value networks on the latents rolled out by the dynamics model, rather than only the directly encoded latents. For a given trajectory in the replay buffer:
\[
z_t, a_t, r_t, ..., z_{t+H-1}, a_{t+H-1}, r_{t+H-1}, z_{t+H}
\]
the dynamics head rolls out the same actions $[a_t, ..., a_{t+H-1}]$ from the start state $z_t$ to create the estimated latents:
\begin{equation}
\hat{z}_{t+1}, \ldots, \hat{z}_{t+H} \quad \text{where} \quad \hat{z}_{t+u+1} = f_{\theta}(\hat{z}_{t+u}, a_{t+u}).
\end{equation}
These estimated latents are not only used to train the dynamics head, but also to train the reward and value networks. Specifically, the reward head is trained to predict the \textit{true} reward $r_{t+u}$, using the estimated latent $\hat{z}_{t+u}$. Similarly, the value head is trained to predict the value target corresponding to the \textit{true} latent $y(z_{t+u})$, using only the estimated latent $\hat{z}_{t+u}$. The value target $y(z_{t+u})$ corresponding to the true latent $z_{t+u}$ is calculated as $\hat{Q}^\pi_1(z_{t+u},a_{t+u} \sim \pi_\theta(z_{t+u}))$.
The result of training networks on the rolled out latents is that the value/reward networks become coupled to the dynamics model. If the dynamics model makes a large error, the value/reward networks will be trained to compensate for it.




\textbf{Planner distillation and reanalyze:} During acting, BMPC stores $\pi_{\mathrm{plan}}$ as a policy target and trains $\pi_\theta$ with $D_{\mathrm{KL}}(\pi_\theta(\cdot \mid z_t)\,\|\,\pi_{\mathrm{plan}})$. However, these targets become stale over time, so they are reanalyzed \citep{schrittwieser2020muzero} by replanning on replay states. Since these refreshed policy targets are never executed in the environment, there is no new data collected to ensure the model and value heads are accurate on them.

\section{Related work}
\label{sec:related-work}

\textbf{TD-MPC family and planner distillation}
\label{sec:rw-tdmpc-family}
As discussed in Section \ref{sec:background}, reanalyzed planner targets distilled into the policy are never executed in the environment; therefore errors exploited in those targets may never be corrected through interaction with the environment.
BOOM \citep{zhan2025boom} and TD-M(PC)$^2$ \citep{lin2025tdmpc-squared} address this issue by regularizing the policy towards the action taken in the environment, which leads to large improvements on tasks with high-dimensional action spaces. A downside, however, is that the policy is regularized towards stale actions, which is suboptimal. BMPC \citep{wang2025bmpc} then suggests to train the policy fully with distillation, using reanalyze \citep{schrittwieser2020muzero} to refresh the stale policy targets. 

\textbf{Multi-horizon value estimation}
\label{sec:rw-horizon}
Combining value estimates across multiple rollout depths is a common strategy in (model-based) RL to reduce the variance of a return estimate.
DreamerV3 \citep{hafner2025dreamerv3} rolls out policy actions in its learned model and uses TD($\lambda$) \citep{sutton1988td} to combine the return estimates from different rollout depths, yielding lower variance value estimates.
MCTS-based methods like TSMCTS \citep{oren2025tsmcts} aggregate the value of child nodes at different depths to reduce the variance of their value estimates. To our knowledge, combining return estimates across multiple rollout depths has not been done for MPC objectives.

\textbf{Dynamics ensembles in model-based reinforcement learning}
\label{sec:rw-ensembles}
Ensembles of learned dynamics models are widely used in MBRL. PETS \citep{chua2018pets}, MBPO \citep{janner2019mbpo}, MOPO \citep{yu2020mopo} and STEVE \citep{buckman2018steve} use bootstrapped dynamics ensembles for uncertainty-aware model rollouts. The ensembles reduce variance in the rollouts, but also provide an estimate of the epistemic uncertainty of the model rollout. To our knowledge, none of these methods are value-equivalent/MuZero-style, meaning none train the value/reward networks on the latents rolled out by multiple dynamics models (See section \ref{sec:background}).

\textbf{Pessimism in reinforcement learning}
\label{sec:rw-pessimism}
Pessimism in reinforcement learning is a strategy to avoid epistemically uncertain regions \citep{rashidinejad2023pessimisticrl}. Pessimism is often implemented as a penalty to the value estimates, either by taking a minimum over an ensemble or subset of the ensemble \citep{chen2021redq,haarnoja2019sac,oren2024viac,hansen2024tdmpc2,fujimoto2018td3} or by subtracting a disagreement based penalty \citep{yu2020mopo}. This penalty can be applied to value targets \citep{haarnoja2019sac,chen2021redq,fujimoto2018td3}, to the policy objective \citep{mani2026shapo}, or to the planner objective \citep{chua2018pets, janner2019mbpo, yu2020mopo, oren2025emcts}. Mrs.\ Q \citep{mrsq2026} recently showed that pessimism is critical for achieving strong performance on HumanoidBench~\citep{sferrazza2024humanoidbench}, but also notes that it can hurt performance on DMC~\citep{tassa2018dmcontrol}.

\section{EfficientTDMPC}
\label{sec:method}
We introduce EfficientTDMPC, a TD-MPC-style method that improves the reliability of model-based return estimates used for planning.
The TD-MPC family relies heavily on its planner to find actions that maximize the estimated return. Reducing the error in the return estimate can be expected to yield better results from planning.
The return estimate is obtained by rolling out the learned dynamics and reward models, and bootstrapping with the value network at the end of the rollout. As such, the return estimate is prone to error from all components, which compounds with rollout depth. EfficientTDMPC makes the planner objective less sensitive to errors in any single learned model rollout.

First, it uses a multi-horizon planning objective that averages the return estimates from rolling out and bootstrapping at different depths, which has been successful in other works that employ MCTS \citep{browne2012mcts, silver2017alphazero, schrittwieser2020muzero}, and is known to reduce estimator variance under weak assumptions \citep{oren2025tsmcts, browne2012mcts}.
Secondly, it introduces a value-equivalent dynamics ensemble, whose return predictions are averaged, which previous work has shown can reduce error \citep{chua2018pets, janner2019mbpo, yu2020mopo} of model-based return estimates. 
Third, it introduces pessimistic reanalyze, which prevents distilling action distributions into the policy for which the value estimates are very uncertain. We then introduce two practical training improvements that improve replay buffer freshness and reduce wall-clock training time. We build our contributions on BMPC \citep{wang2025bmpc}, a recent strong contribution to the TD-MPC family.

\begin{figure}[H] 
	\centering
	\includegraphics[width=\linewidth]{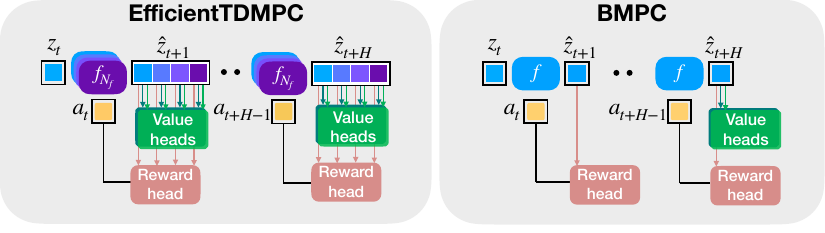}
	\caption{\textbf{Process of return estimation of EfficientTDMPC vs BMPC} (Right) BMPC rolls out a single dynamics head, predicts the reward earned at each depth and bootstraps with its value ensemble. (Left) EfficientTDMPC creates multiple rollouts using an ensemble of dynamics models. It then predicts the reward and bootstrapped value at each depth. The estimate is then averaged over the different rollouts and the return estimates from different rollout depths.}
	\label{fig:method-overview}
\end{figure}

\subsection{Multi-horizon planning objective}
\label{sec:aggregate}
We introduce a multi-horizon planning objective, obtained by averaging the state-action value estimates across different rollout horizons.
As discussed in Section \ref{sec:related-work}, aggregating model-based return estimates across different rollout horizons has been a successful way to reduce estimator variance \citep{sutton1988td, feinberg2018mve, buckman2018steve}. 
However, these prior works roll out action sequences from the \textit{policy network}, so each depth is an estimate of the same quantity, the value. As such, they are simply averaging estimates of the same quantity to get a reduced variance estimate. 
In our work, the action sequence is proposed by the MPC planner rather than by the policy network. Bootstrapping with the value function at depth $h$ replaces every later planned action with the policy, so the expected return changes with $h$. Each term is the value of a different prefix of the planned sequence, not another estimate of the same value. We average these prefix values:
\begin{equation}
	\hat{Q}_{\mathrm{aggregate},H}(z_t,\mathbf{a}_{t:t+H-1})
	=
	\frac{1}{H}
	\sum_{h=1}^H
	\hat{Q}_h(z_t, \mathbf{a}_{t:t+h-1})
	\label{eq:aggregate-horizon-objective}
\end{equation}
where each $\hat{Q}_h$ only evaluates the value of the first $h$ actions in the action sequence, and bootstraps with the value network at depth $h$. 
The objective favors action sequences whose prefixes remain valuable under policy continuation, rather than trajectories whose value only becomes high at the terminal planning depth.
We use the multi-horizon planning objective during planning at acting time, together with an increase in the rollout horizon.

\subsection{Value-equivalent dynamics ensemble}
\label{sec:dynamics-ensemble}

To the best of our knowledge, we are the first to propose to train a value-equivalent dynamics ensemble. We then use this ensemble to create improved value estimates during planning and value target creation.
Prior work has shown that dynamics ensembles can reduce the error in model-based return estimates \citep{chua2018pets}, but, to the best of our knowledge, this has only been shown for methods that train the dynamics separately from the value and reward networks. 
We jointly train an ensemble of dynamics heads and choose to train the value and reward networks on the latents rolled out by \textit{all} dynamics heads.
We will denote the dynamics heads as $f_{\theta,i}$ for $i \in \{1, .., N_f\}$, where $N_f$ is the number of dynamics heads.

\textbf{Training} We initialize heads differently and train each dynamics head on the same sampled trajectories from the replay buffer ($z_t, a_t, z_{t+1}, ..., z_{t+H-1}, a_{t+H-1}, z_{t+H}$), each then rolls out the action sequence $(a_t, ..., a_{t+H-1})$ from the initial state $z_t$ to create its estimated latents $\hat{z}^i_{t+u+1} = f_{\theta,i}(\hat{z}^i_{t+u}, a_{t+u})$. We then train the reward and value heads to use the estimated latents $\hat{z}^i_{t+u}$ from each dynamics head to predict the true reward $r_{t+u}$ and value target $y(z_{t+u})$ corresponding to the true latent. This implies that the value and reward heads are trained to compensate for the errors of multiple dynamics models.
In Section \ref{sec:exp-ablations} we show that introducing an ensemble of dynamics heads into this system can significantly improve sample efficiency. Details in Appendix \ref{app:training-losses}.

\textbf{Value estimation} 
We use this ensemble for improved model based return estimates, which are used during planning and for creating value targets. 
Similar to $TS_\infty$ sampling from PETS \citep{chua2018pets}, we roll out action (sequences) with a single dynamics head, estimate the value and then use the mean and variance over the ensemble.
We define a state-action value estimate that uses a single dynamics head with index $i \in \{1, .., N_f\}$ and a single value head with index $j \in \{1, .., N_v\}$ as,
\begin{align}
	\hat{Q}_H^\pi(z_t, \mathbf{a}_{t:t+H-1}; i, j) 
\;=\;
\sum_{u=0}^{H-1}\gamma^u R_{\theta}(\hat{z}_{t+u}, a_{t+u})
\;+\;
\gamma^H V_{\theta,j}(\hat{z}_{t+H}),
\,\, \hat{z}_{t+u+1} = f_{\theta,i}(\hat{z}_{t+u}, a_{t+u})
\end{align}
where $\hat{z}_t = z_t$. This is the state-action value estimate obtained by rolling out the dynamics head with index $i$ and bootstrapping with the value head with index $j$. We can then obtain $N_f \times N_v$ different estimates of the state-action value by varying $i$ and $j$. 
We then estimate the mean of the state-action value and its variance,
\begin{align}
	\hat{Q}_H^\pi(z_t, \mathbf{a}_{t:t+H-1})
	=
	\mathbb{E}_{i,j}[\hat{Q}_H^\pi(z_t, \mathbf{a}_{t:t+H-1}; i, j)]
\end{align}
\begin{align}
	\hat{\sigma}^2_{\hat{Q}_H^\pi}(z_t, \mathbf{a}_{t:t+H-1})
=
\frac{Var_{i,j}[\hat{Q}_H^\pi(z_t, \mathbf{a}_{t:t+H-1}; i, j)]}{N_f N_v}
\end{align}	

This variance of the mean estimate, written as $\hat{\sigma}^2_{\hat{Q}_H^\pi}$, is then used in Section \ref{sec:pessimistic-reanalyze} as a penalty to the planner objective during reanalyze. We acknowledge that this estimate underestimates the true variance because it assumes that the individual estimates are independent, which is not the case since the same value or dynamics head is used for several of them. In practice we find this is not an issue.

\subsection{Pessimistic reanalyze}
\label{sec:pessimistic-reanalyze}
We propose pessimistic reanalyze, which adds an optional penalty to the planner objective during reanalyze, to avoid distilling actions into the policy on which the model and value networks are not well trained. 
As discussed in Section \ref{sec:related-work}, a known issue in the TD-MPC family is that of a mismatch between the actions taken in the environment and the actions output by the policy network \citep{zhan2025boom}. 
Since the model and value networks are not trained on the policy actions, they are more likely to produce inaccurate predictions for them. 
Prior work solved this by regularizing the policy towards the action that was taken in the environment, but in later parts of the training that action becomes stale, which can be expected to slow down learning. BMPC solves this by reanalyzing the stale targets, but by doing so it reintroduces the issue where the model may not be trained on the actions output by the policy.
EfficientTDMPC introduces pessimistic reanalyze to make policy targets that are fresh, but are also regularized towards actions that the model has actually been trained on. Specifically, we define the pessimistic reanalyze objective as follows:
\begin{align}
J_{\mathrm{reanalyze},H}(\mathbf{a}_{t:t+H-1}) = \hat{Q}_H^\pi(z_t, \mathbf{a}_{t:t+H-1}) - \beta \hat{\sigma}_{\hat{Q}_H^\pi}(z_t, \mathbf{a}_{t:t+H-1}),
\end{align}
where $\hat{\sigma}_{\hat{Q}_H^\pi}(z_t, \mathbf{a}_{t:t+H-1})$ is the square root of the estimated variance defined in Section \ref{sec:dynamics-ensemble}. Here $\beta$ trades off maximizing expected return and minimizing uncertainty. A higher $\beta$ leads to a planner that finds actions with more certain return estimates, which are more likely to be accurate. Since the reanalyze targets are distilled into the policy, the policy also outputs actions that lead to more certain return estimates, which in turn leads to more reliable value targets. We note that the option to add pessimism is fundamentally a tradeoff that different tasks respond differently to. We find in Section \ref{sec:exp-ablations} that a useful heuristic is to apply pessimism to tasks that have early termination.

\subsection{Practical modifications}
\label{sec:practical}

 \textbf{Data freshness}
EfficientTDMPC inserts each transition immediately, as opposed to previous work in the TD-MPC family which adds newly collected transitions after a full episode. This shortens the delay before new experience can affect model, value, and policy learning.

\textbf{Cheap reanalyze planning}
Reanalyze is responsible for the majority of the wall-clock training time, so we reduce its planning budget by reducing the particles, elites and policy seeds by a factor of eight. We find in Section \ref{sec:exp-ablations} that, for reanalyze, fewer planning particles are sufficient in practice. 


\section{Experiments}
\label{sec:experiments}

We implement our EfficientTDMPC agent using a value-equivalent dynamics ensemble with four dynamics heads. During planning at acting time we increase the rollout horizon from three to six and optimize the multi-horizon planning objective. To save compute we do not apply the multi-horizon planning objective to planning during reanalyze. To decrease compute further we reduce the number of reanalyze planning particles, elites and policy seeds by a factor of 8.
We find that pessimistic reanalyze is beneficial on some tasks and harmful on others.
The pattern is consistent with early episode termination (tasks that can terminate early tend to benefit; others do not), in line with prior work \citep{mrsq2026,oren2024viac,mani2026shapo}.
We therefore apply pessimistic reanalyze with a fixed $\beta=10$ only on tasks that can terminate early.
We evaluate on the 7 hard and 21 easy DMC tasks as well as the 14 commonly used HumanoidBench with hand tasks.
We describe the baselines used for comparison in Appendix~\ref{app:baseline-details}.

\subsection{Results}
\label{sec:results}
\vspace{-0em}
\begin{figure}[H]
    \centering
    \vspace{-1.0em}
    \includegraphics{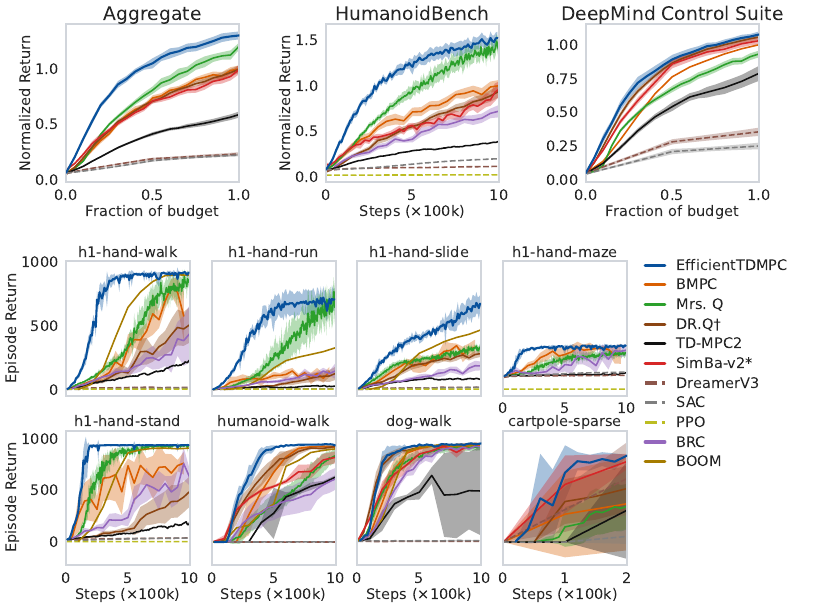}
    \vspace{-1.2em}
    \caption{\textbf{Learning curves.} Top, left to right: equal-weight aggregate (0.5 HumanoidBench + 0.5 DeepMind Control Suite), HumanoidBench (14 tasks), and the DeepMind Control Suite (suite-equal Easy and Hard). Bottom: learning curves on selected tasks. Shaded regions are 95\% CIs. $^*$SimBa-v2 is evaluated on HumanoidBench \emph{without} hands. $^\dagger$DR.Q is missing several HumanoidBench tasks. Full per-task grids are in Appendices~\ref{app:pertaskhb},~\ref{app:pertaskharddmc}, and~\ref{app:pertaskeasydmc}.}
    \label{fig:aggregate-learning-curves}
\end{figure}
\noindent Figure~\ref{fig:aggregate-learning-curves} shows an equal-weight aggregate of HumanoidBench and the DeepMind Control Suite, the two suites separately, and selected tasks where EfficientTDMPC substantially outperforms prior methods, reaching BMPC's final aggregate performance using 57\% fewer environment steps. EfficientTDMPC is state of the art in terms of sample efficiency on HumanoidBench and the DeepMind Control Suite.

\subsection{Component ablations}
\label{sec:exp-ablations}

\begin{figure}[H]
    \centering
    \includegraphics{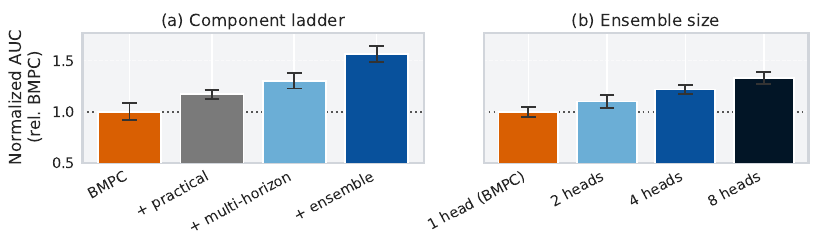}
    \caption{\textbf{Contribution ablation.}
    (\subref{fig:component-ladder}) Cumulative component ladder adding practical changes, then the multi-horizon planning objective, then the value-equivalent dynamics ensemble (4 dynamics heads). We show normalized AUC over 5 tasks (quadruped-walk, reacher-hard, dog-stand, humanoid-walk, h1hand-walk) using 5+ seeds each.
    (\subref{fig:dynamics-heads}) Adding a value-equivalent dynamics ensemble of different sizes to BMPC.}
    \label{fig:component-ladder-heads}
    \phantomsubcaption\label{fig:component-ladder}
    \phantomsubcaption\label{fig:dynamics-heads}
\end{figure}

\textbf{Contribution ladder and ensemble scaling.}
We add our changes to BMPC cumulatively and measure normalized AUC on five representative tasks (quadruped-walk, reacher-hard, dog-stand, humanoid-walk, h1hand-walk; 5--8 seeds).
Figure~\ref{fig:component-ladder-heads}\subref{fig:component-ladder} shows the result.
Our practical improvements (horizon increase, per-step buffer insertion, cheaper reanalyze) already mildly improve performance; adding the multi-horizon planning objective and the value-equivalent dynamics ensemble then yields further gains, for a final $+56\%$ AUC relative to BMPC.
Figure~\ref{fig:component-ladder-heads}\subref{fig:dynamics-heads} isolates the effect of adding a value-equivalent dynamics ensemble of different sizes to BMPC. We find the AUC on the ablation tasks  increases by 10\%, 22\%, and 33\% for 2, 4, and 8 dynamics heads, respectively. Note that EfficientTDMPC uses only 4 dynamics heads to limit wall-clock training time.

\begin{figure}[H]
    \centering
    \includegraphics{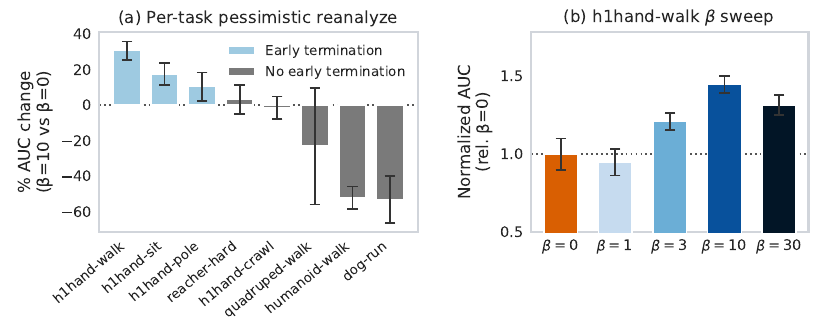}
    \caption{\textbf{When to use pessimistic reanalyze, and which $\beta$.}
    (\subref{fig:pessimism-delta}) Percent AUC change from adding pessimistic reanalyze ($\beta{=}10$ vs.\ $\beta{=}0$) on h1hand-walk, h1hand-sit, h1hand-pole, h1hand-crawl, reacher-hard, quadruped-walk, dog-run, and humanoid-walk (8 seeds; 95\% CIs). Light blue: early termination; grey: no early termination.
    (\subref{fig:pessimism-beta-sweep}) h1hand-walk normalized AUC for $\beta \in \{0,1,3,10,30\}$ (6 seeds; 95\% CI); $\beta{=}10$ is our default.}
    \label{fig:pessimism-breadth}
    \phantomsubcaption\label{fig:pessimism-delta}
    \phantomsubcaption\label{fig:pessimism-beta-sweep}
\end{figure}

\textbf{Pessimistic reanalyze.}
Figure~\ref{fig:pessimism-breadth} analyzes the effect of applying pessimistic reanalyze.

Panel~(\subref{fig:pessimism-delta}) reports the percentage change in AUC by adding pessimistic reanalyze ($\beta{=}10$ vs.\ $\beta{=}0$) on 8 tasks (four HumanoidBench, four DMC). In line with prior work \citep{mrsq2026,oren2024viac,mani2026shapo}, we find that tasks that terminate early benefit from pessimistic reanalyze, while tasks that do not, such as DMC tasks and h1hand-crawl, do not benefit. For this reason, we apply pessimistic reanalyze only on tasks that can terminate early.
Panel~(\subref{fig:pessimism-beta-sweep}) reports normalized AUC on h1hand-walk for $\beta \in \{0,1,3,10,30\}$, where pessimistic reanalyze helps. Moderate $\beta$ improves over $\beta{=}0$, $\beta{=}10$ is best, and $\beta{=}30$ is worse than $\beta{=}10$ but still above $\beta{=}0$.

\begin{figure}[H]
    \centering
    \includegraphics{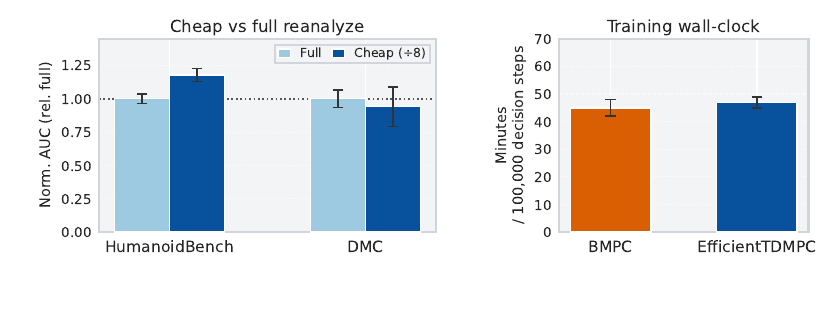}
    \caption{\textbf{Cheap reanalyze and wall-clock.} Left: normalized AUC of cheap reanalyze (64 particles, 8 elites, and 3 policy seeds; $\div 8$ budget) relative to full reanalyze (512). HumanoidBench: four h1hand tasks (5 seeds). DMC: quadruped-walk, reacher-hard, dog-stand, humanoid-walk ($\ge$5 seeds). 95\% CIs. Right: training wall-clock in minutes per 100,000 decision steps on one NVIDIA A100 ($47{\pm}2$ for EfficientTDMPC vs.\ $45{\pm}3$ for BMPC).}
    \label{fig:cheap-reanalyze}
\end{figure}

\textbf{Cheap reanalyze and wall-clock.}
Reanalyze dominates training time, so we reduce its planning budget by a factor of eight (particles, elites, and policy seeds).
Figure~\ref{fig:cheap-reanalyze} shows that this slightly \emph{improves} HumanoidBench AUC and only slightly reduces DMC, while keeping wall-clock training time similar to BMPC.

\section{Conclusion and limitations}
\label{sec:conclusion}

EfficientTDMPC introduces three improvements to MPC objectives in model-based RL. We apply these improvements to a baseline in the TD-MPC family and show that each significantly increases sample efficiency.
Firstly, we replace the standard value bootstrapped objective with our multi-horizon planning objective, which requires the estimated value to be high along the full rollout, not just at the end. 
Secondly, we average the objective over rollouts from multiple dynamics heads. To the best of our knowledge, we are the first to train a value-equivalent dynamics ensemble. 
Thirdly, we introduce the option of pessimistic reanalyze to the TD-MPC framework, which aims to discourage the policy network from exploiting estimation errors. We find that this is beneficial in environments that can terminate early, so we apply it to those only. We do not yet show that this heuristic is effective on other benchmarks.
The dynamics ensemble adds additional compute, but this can be compensated for by making reanalyze planning cheaper, which we find does not significantly hurt performance.
To the best of our knowledge, the resulting EfficientTDMPC is state of the art in sample efficiency on HumanoidBench and the DeepMind Control Suite. It also outperforms the baseline's aggregate performance using 57$\%$ fewer environment steps.

\bibliographystyle{plainnat}
\bibliography{references}


\appendix
\newpage
\section{Implementation details}
\label{app:implementation}

Throughout this paper, HumanoidBench refers to the 14-task HumanoidBench suite with hands (the H1 hand embodiment), not the version without hands.

\subsection{Hyperparameters}
\label{app:hyperparameters}

Table~\ref{tab:shared-hyperparams} compares the default EfficientTDMPC configuration used in the main paper against base BMPC. Entries in the EfficientTDMPC column are boldfaced when they differ from BMPC.

\begin{table}[H]
\centering
\caption{\textbf{EfficientTDMPC versus BMPC hyperparameters.} Bold entries in the EfficientTDMPC column denote values that differ from BMPC. The main-paper agent uses one model/value/policy update per environment step (very similar wall-clock to BMPC).}
\label{tab:shared-hyperparams}
\small
\begin{tabular}{p{2.3cm}p{3.5cm}p{2.7cm}p{3cm}}
\toprule
\textbf{Category} & \textbf{Parameter} & \textbf{EfficientTDMPC} & \textbf{BMPC} \\
\midrule
\multirow{8}{*}{Architecture}
  & Dynamics heads       & \textbf{4} & 1 \\
  & Latent dim           & 512 & 512 \\
  & MLP hidden dim       & 512 & 512 \\
  & MLP layers           & 2 & 2 \\
  & Encoder dim / layers & 256 / 2 & 256 / 2 \\
  & SimNorm dim          & 8 & 8 \\
  & Value heads          & 2 & 2 \\
  & Reward heads         & 1 & 1 \\
\midrule
\multirow{7}{*}{Training}
  & Batch size           & 256 & 256 \\
  & Learning rate        & $3 \times 10^{-4}$ & $3 \times 10^{-4}$ \\
  & Encoder LR scale     & 0.3 & 0.3 \\
  & Grad clip norm       & 20 & 20 \\
  & EMA $\tau$           & 0.01 & 0.01 \\
  & Model / Value / Policy updates per env step & 1 / 1 / 1 & 1 / 1 / 1 \\
  & Reanalyze interval   & 10 & 10 \\
\midrule
\multirow{8}{*}{\shortstack{Acting planner\\(MPPI)}}
  & Horizon              & \textbf{6} & 3 \\
  & Multi-horizon planning objective  & \textbf{mean} & final \\
  & Iterations           & 6 (+2 if $|A| > 20$) & 6 (+2 if $|A| > 20$) \\
  & Samples              & 512 & 512 \\
  & Elites               & 64 & 64 \\
  & Policy trajectories  & 24 & 24 \\
  & Temperature          & 0.5 & 0.5 \\
  & Min / Max std        & 0.05 / 2.0 & 0.05 / 2.0 \\
\midrule
\multirow{10}{*}{\shortstack{Reanalyze\\planner (MPPI)}}
  & Horizon              & 3 & 3 \\
  & Multi-horizon objective  & final & final \\
  & Iterations           & 6 (+2 if $|A| > 20$) & 6 (+2 if $|A| > 20$) \\
  & Samples              & \textbf{64} & 512 \\
  & Elites               & \textbf{8} & 64 \\
  & Policy trajectories  & \textbf{3} & 24 \\
  & Temperature          & 0.5 & 0.5 \\
  & Min / Max std        & 0.05 / 2.0 & 0.05 / 2.0 \\
  & Batch size           & 20 & 20 \\
  & Pessimistic reanalyze $\beta$     & $\boldsymbol{\beta}=10$ \textbf{(early-term.\ tasks) /} $0$ \textbf{else} & $\beta{=}0$ \\
\midrule
\multirow{3}{*}{Policy}
  & Log std bounds       & $[-3, 1]$ & $[-3, 1]$ \\
  & Entropy coeff        & $10^{-4}$ & $10^{-4}$ \\
  & Optimization         & distillation & distillation \\
\midrule
\multirow{3}{*}{Discount}
  & $\gamma_{\min}$ / $\gamma_{\max}$ & 0.95 / 0.995 & 0.95 / 0.995 \\
  & $\rho$               & 0.5 & 0.5 \\
  & Value bins / range   & 101 / $[-10, 10]$ & 101 / $[-10, 10]$ \\
\midrule
\multirow{1}{*}{Replay}
  & Buffer update interval & \textbf{every step} & every episode \\
\bottomrule
\end{tabular}
\end{table}

The acting planner uses horizon 6 with the multi-horizon planning objective (mean over depths), while the reanalyze planner is configured separately with horizon 3, a terminal-only objective, and the smaller 64 / 8 / 3 sample, elite, and policy budget shown above. In code, both planners receive an additional two iterations automatically when the action dimension is greater than 20. Reanalyze uses batch size 20 for both methods and applies pessimistic reanalyze with $\beta{=}10$ only on tasks that can terminate early (otherwise $\beta{=}0$).

\subsection{Generating performance figures}
\label{app:generating-performance-figures}

We report evaluation return versus environment steps.
HumanoidBench is the 14-task suite with hands (h1hand), trained for 1M steps.
On the DeepMind Control Suite we follow the standard split of 21 Easy tasks (200k steps, as in EfficientZero V2 \citep{wang2024efficientzerov2}) and 7 Hard tasks (1M steps).
EfficientTDMPC main-paper runs use 4 seeds per task unless a caption says otherwise.
Per-task curves for every task are in Appendices~\ref{app:pertaskhb}--\ref{app:pertaskeasydmc}.

\paragraph{Per-task curves.}
Each point is the mean return over seeds.
Shaded regions are 95\% confidence intervals ($\mathrm{mean}\pm 1.96\,\mathrm{SE}$).

\paragraph{Normalization.}
To compare across tasks we divide each task's return by BMPC's mean return on that task at the same budget, so BMPC ends at 1.

\paragraph{Aggregates.}
Within HumanoidBench, and within Easy or Hard DMC, we average the normalized curves with equal weight per task.
The DeepMind Control Suite panel in Figure~\ref{fig:aggregate-learning-curves} then averages the Easy and Hard suite means with equal weight, after aligning each task to the fraction of its own budget.

\paragraph{AUC and Figure~\ref{fig:intro_auc}.}
We interpolate every curve onto a shared step grid, anchoring a missing step~0 at BMPC's untrained score, and take the area under the normalized curve.
We compute this separately on HumanoidBench, Easy DMC, and Hard DMC.
Suite-equal DMC is the mean of the Easy and Hard AUCs.
To highlight EfficientTDMPC's dominance specifically in early environment steps, the introduction bar restricts each of these areas to the first half of that suite's full evaluation budget  (100k of 200k on Easy DMC; 500k of 1M on Hard DMC and HumanoidBench). It averages that HumanoidBench AUC with the suite-equal DMC AUC, then divides by BMPC's AUC over the same half so that BMPC equals 1.
$^*$SimBa-v2 HumanoidBench numbers use the embodiment without hands, scored against the harder BMPC with hands, and are therefore an upper bound.
$^\dagger$DR.Q is missing hurdle, maze, balance\_simple, and balance\_hard on HumanoidBench.

\subsection{Baseline details}
\label{app:baseline-details}

We compare EfficientTDMPC against recent model-based and model-free baselines for continuous control. Whenever possible, we use official result files released by the authors. When official results are not available for a specific benchmark setting, we run the official implementation with the benchmark-specific configuration described below.

\textbf{SAC.}
Soft Actor-Critic (SAC) \citep{haarnoja2019sac} is an off-policy model-free actor-critic baseline. We use the SAC results provided in the official BMPC repository:
\url{https://github.com/wertyuilife2/bmpc}.

\textbf{DreamerV3.}
DreamerV3 \citep{hafner2025dreamerv3} is a latent world-model agent trained with imagination-based policy optimization. We use the DreamerV3 results provided in the official BMPC repository:
\url{https://github.com/wertyuilife2/bmpc}.

\textbf{TD-MPC2.}
TD-MPC2 \citep{hansen2024tdmpc2} is the main TD-MPC-family baseline preceding BMPC, using latent dynamics, value bootstrapping, and MPPI planning. We use the TD-MPC2 results provided in the official BMPC repository:
\url{https://github.com/wertyuilife2/bmpc}.

\textbf{BMPC.}
Bootstrapped Model Predictive Control (BMPC) \citep{wang2025bmpc} is the primary baseline for our method. EfficientTDMPC is built on top of BMPC. We use the official BMPC implementation and released result files from:
\url{https://github.com/wertyuilife2/bmpc}.

\textbf{BOOM.}
BOOM \citep{zhan2025boom} is a model-based RL method that bootstraps an off-policy policy using world-model planning. We use the released BOOM result files from the official BOOM repository:
\url{https://github.com/molumitu/BOOM_MBRL}.

\textbf{BRC.}
Bigger, Regularized, Categorical (BRC) \citep{nauman2025BRC} is a strong model-free baseline based on high-capacity categorical value functions. We use the released result files from the official BRC repository:
\url{https://github.com/naumix/BiggerRegularizedCategorical/tree/main/results}.

\textbf{SimBa-v2.}
SimBa-v2 \citep{lee2025simbav2} is a strong model-free baseline based on hyperspherical normalization. For DMC, we use the released result files from the official SimBa-v2 repository:
\url{https://github.com/DAVIAN-Robotics/SimbaV2/tree/master/results}.
For HumanoidBench, official SimBa-v2 results use the H1 embodiment \emph{without} hands. That is a simpler special case of HumanoidBench (with hands); we therefore treat those SimBa-v2 numbers as an upper bound. We run the official implementation with action repeat 1, matching the HumanoidBench protocol.

\textbf{Mrs.\ Q.}
Mrs.\ Q \citep{mrsq2026} is a recent baseline that studies pessimism and search in model-based reinforcement learning. We use the released result files from the official Mrs.\ Q repository:
\url{https://github.com/facebookresearch/MRSQ/tree/main/results}.

\textbf{DR.Q.}
DR.Q \citep{lyu2026drq} is a recent model-free baseline that learns debiased model-based representations for Q-learning. We use the authors' released result files:
\url{https://github.com/dmksjfl/DR.Q}.
On HumanoidBench, DR.Q is missing several tasks (\texttt{hurdle}, \texttt{maze}, \texttt{balance\_simple}, \texttt{balance\_hard}).

\textbf{PPO.}
Proximal Policy Optimization (PPO) \citep{schulman2017ppo} is included as an additional model-free baseline on HumanoidBench. We use the PPO results from the official HumanoidBench codebase:
\url{https://github.com/carlosferrazza/humanoid-bench}.

\textbf{EfficientZero V2.}
EfficientZero V2 \citep{wang2024efficientzerov2} is an extension of MuZero for sample efficiency. 
\url{https://github.com/shengjiewang-jason/efficientzerov2}

\subsection{Wall-clock comparison}
\label{app:runtime-comparison-implementation}
We compare wall-clock training time of EfficientTDMPC against the official BMPC implementation on a single NVIDIA A100 GPU (excluding compilation and evaluation). Runs were timed over 200k environment steps, which with action repeat 2 is 100k decision steps; we report the original measured times as minutes per 100,000 decision steps ($47{\pm}2$ for EfficientTDMPC vs.\ $45{\pm}3$ for BMPC). Both methods use one model, value, and policy update per environment step.
\newpage
\section{Per task results}
\label{app:per-task-results}

\subsection{Per-task learning curves HumanoidBench}
\label{app:pertaskhb}

\vspace*{\fill}
\begin{figure}[H]
    \centering
  \includegraphics{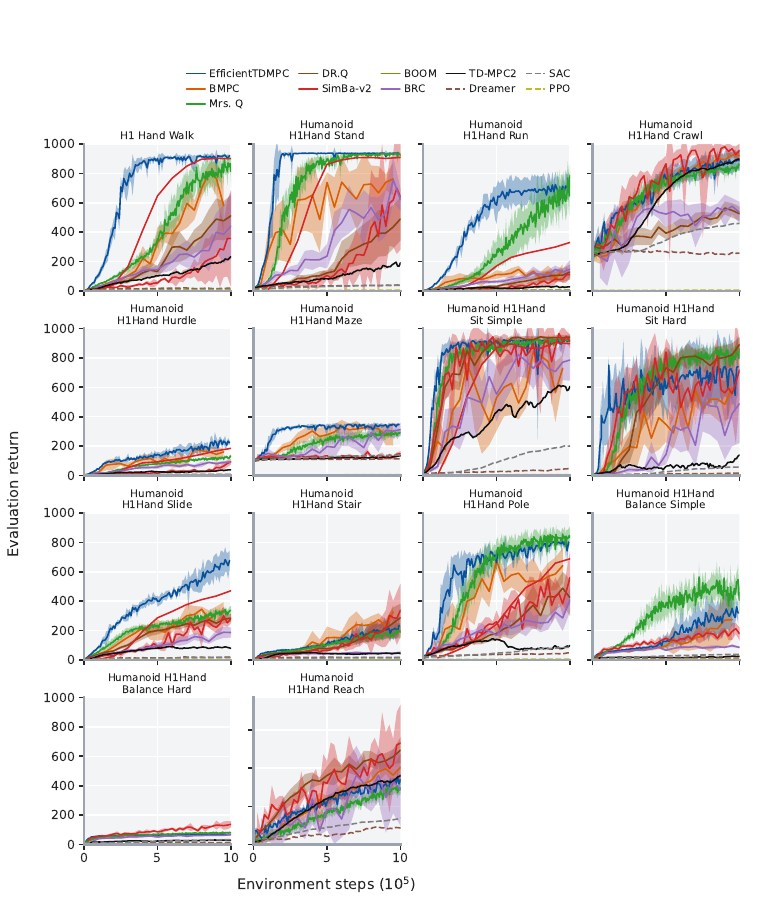}
  \caption{\textbf{Per-task results: HumanoidBench (1M environment steps).} Evaluation return on all 14 h1hand tasks with all available baselines. EfficientTDMPC is the default main-paper agent.}
    \label{fig:app-humanoidbench-tasks}
\end{figure}
\vspace*{\fill}
\clearpage

\subsection{Per-task learning curves Hard DMC}
\label{app:pertaskharddmc}

\vspace*{\fill}
\begin{figure}[H]
    \centering
  \includegraphics{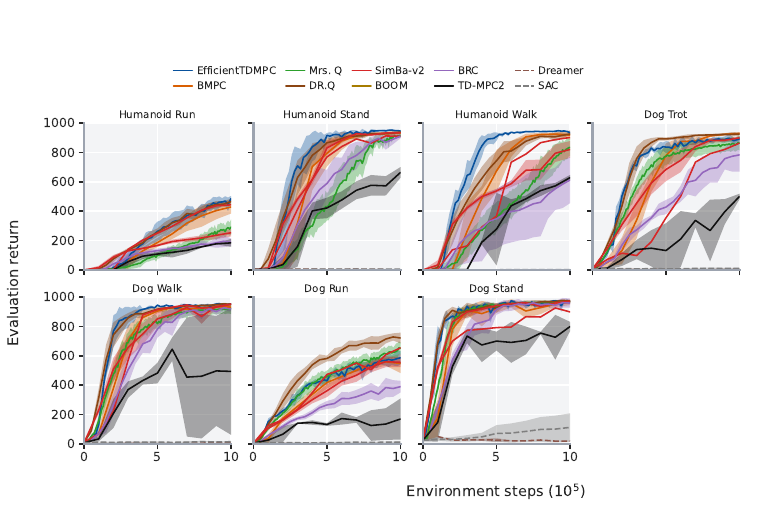}
  \caption{\textbf{Per-task results: Hard DMC (1M environment steps).} Evaluation return on all 7 hard DMC tasks with all available baselines. EfficientTDMPC is the default main-paper agent.}
    \label{fig:app-hard-dmc-1m-env}
\end{figure}
\vspace*{\fill}
\clearpage

\subsection{Per-task learning curves Easy DMC}
\label{app:pertaskeasydmc}

\vspace*{\fill}
\begin{figure}[H]
    \centering
  \includegraphics{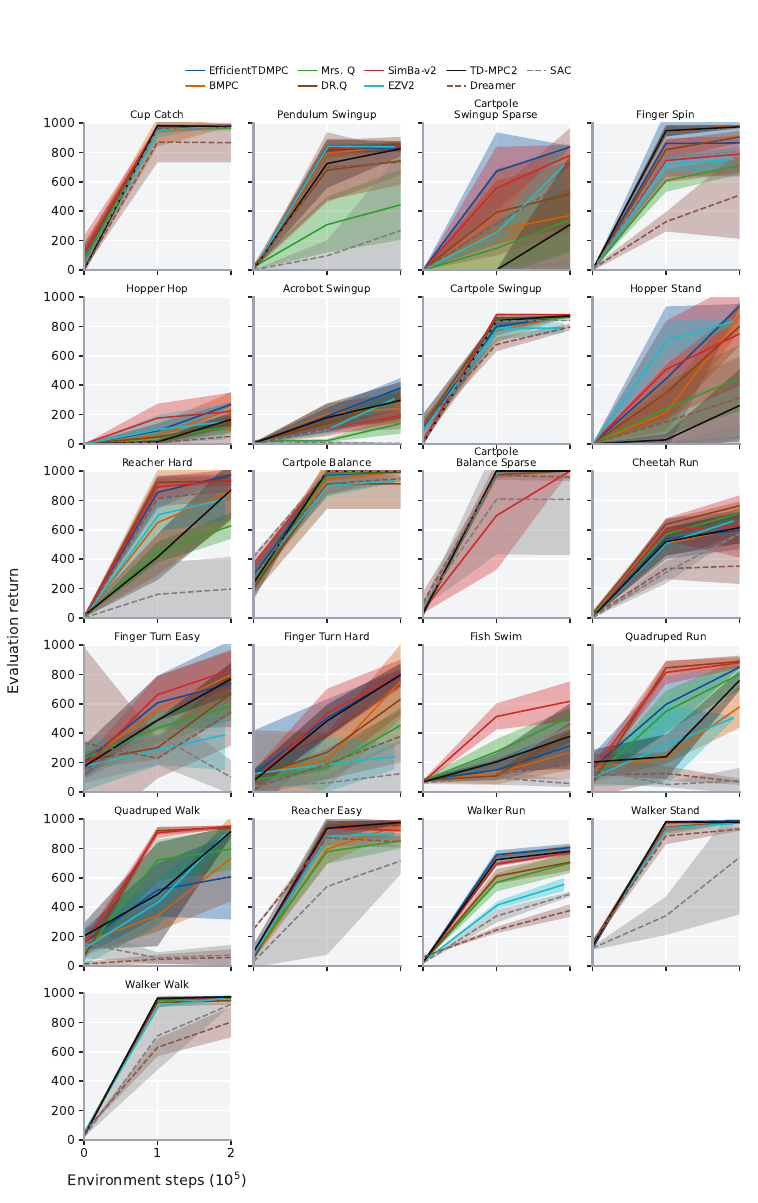}
  \caption{\textbf{Per-task results: Easy DMC (200k environment steps).} Evaluation return on all 21 easy DMC tasks with all available baselines. EfficientTDMPC is the default main-paper agent.}
    \label{fig:app-easy-dmc-200k-env}
\end{figure}
\vspace*{\fill}
\clearpage


\subsection{HumanoidBench pessimistic-reanalyze scope}
\label{app:hb-pessimism-scope}


EfficientTDMPC applies pessimistic reanalyze only. We ablate the effect of applying the same uncertainty penalty when acting in the environment as well. On 4 tasks from HumanoidBench, we compare pessimistic reanalyze against applying the penalty during all planning. We find that additionally applying the penalty during acting does not improve performance. We use the same coefficient $\beta{=}10$ for both conditions.

\begin{figure}[H]

      \centering
      \includegraphics{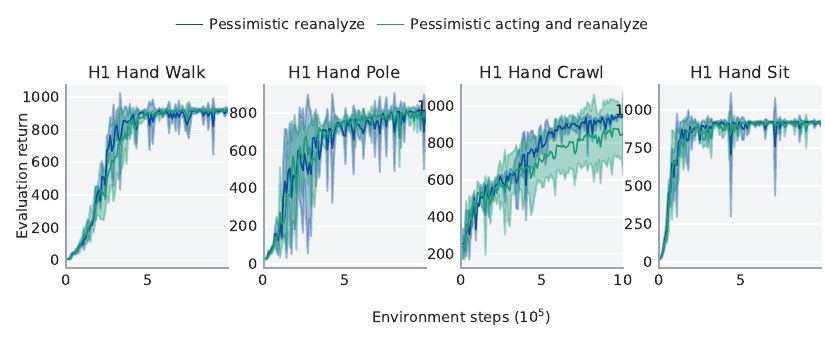}

  \caption{\textbf{HumanoidBench pessimistic-reanalyze scope.} Per-task evaluation return versus environment interactions on four HumanoidBench core tasks when the uncertainty penalty is applied during reanalyze only versus during training-time planning, evaluation-time planning, and reanalyze. }
  \label{fig:app-hb-pessimism-scope}
\end{figure}

\newpage
\section{Qualitative analysis}
\label{app:qualitative}

\subsection{Planner cross-scoring quantities}
\label{app:analysis-quantities}
We study the effect of averaging the planner objective over the ensemble. On four tasks we reload agent and replay-buffer checkpoints from training and plan from replay states. Each candidate action sequence is scored with three return estimates. The single-head estimate uses one value head and one dynamics head. The ensemble estimate averages the two value heads and four dynamics heads, which is the EfficientTDMPC acting objective. The environment estimate rolls the same three actions out in the simulator and bootstraps with the learned value heads, which we treat as a proxy for the true return.

\begin{figure}[H]
    \centering
    \includegraphics{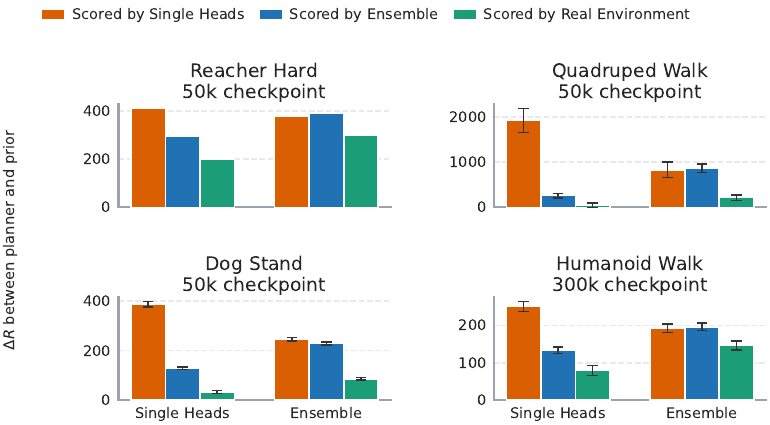}
  \caption{\textbf{Ensemble averaging of return estimate reduces single-head planner exploitation.} The plotted $\Delta R$ quantity estimates how much each planner is predicted to outperform the policy action from the same state. Bars show means over 512 replay states and error bars show standard errors over states.}
    \label{fig:analysis-exploitation}
\end{figure}

Planning is run twice from each state, once under the single-head estimate and once under the ensemble estimate. Both runs use the reduced reanalyze budget: six MPPI iterations, 64 samples, eight elites, three policy trajectories, and temperature 0.5. Every planned sequence, and the policy action from the same state, is then scored under all three estimates. $\Delta R$ is the planner score minus the policy-action score, scaled from the three-step window to a 500-step episode. It estimates how much that planner is predicted to outperform the policy action. Figure~\ref{fig:analysis-exploitation} shows the mean of $\Delta R$ over 512 replay states. Error bars are the standard error across those states.

Sequences maximized under the single-head estimate receive a very high score from that same estimate. Scored in the simulator, their return is much lower and is often worse than the policy action. Sequences maximized under the ensemble estimate are given a smaller advantage over the policy by their own objective, and that advantage agrees more closely with the simulator. Averaging the return over the ensemble therefore makes the planner less able to exploit an error in one value head and one dynamics head.

\subsection{Latent trajectory gallery}
\label{app:latent-trajectory-gallery}


\begin{figure}[H]
    \centering
  \includegraphics{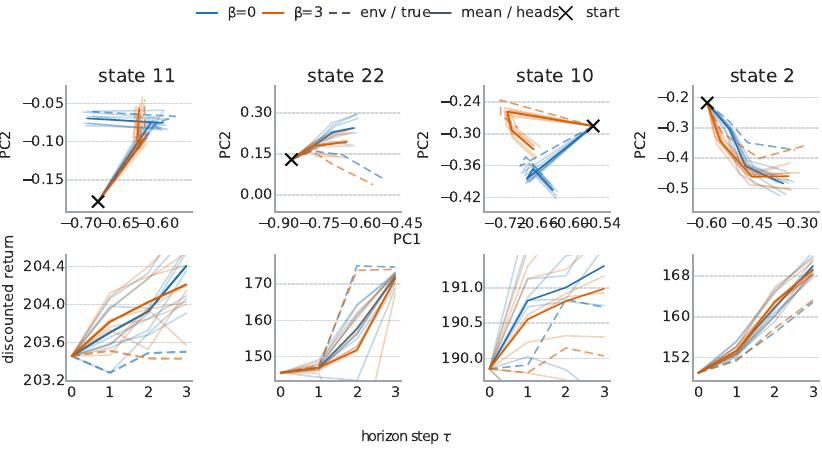}
  \caption{\textbf{Latent trajectory gallery on reacher-hard at 50k (four states).} Blue denotes actions from the optimistic planner ($\beta=0$) and orange from the pessimistic planner ($\beta=3$). The top row shows the latent trajectory in PCA space, both for each head's predicted trajectory, the mean trajectory across heads, and the true environment. The bottom row shows the estimated return of each trajectory at each depth.}
    \label{fig:app-latent-gallery-reacher}
\end{figure}

\begin{figure}[H]
    \centering
  \includegraphics{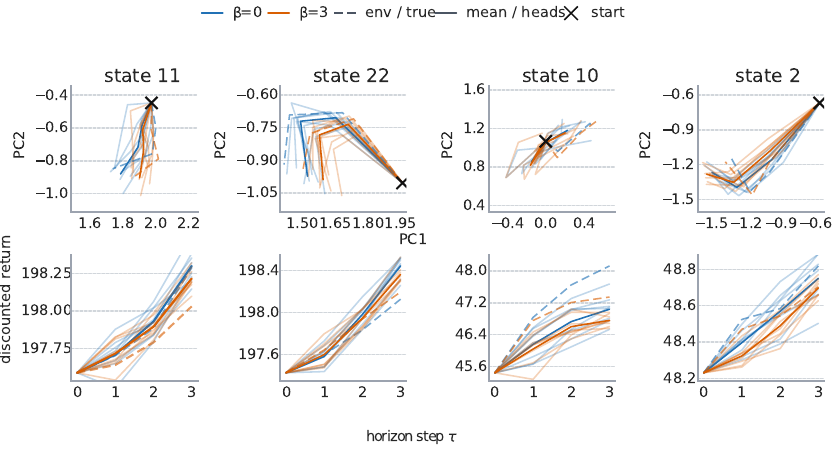}
  \caption{\textbf{Latent trajectory gallery on quadruped-walk at 50k (four states).} Same as Figure~\ref{fig:app-latent-gallery-reacher} but for quadruped-walk. Note that for some states the estimated value increases significantly. }
    \label{fig:app-latent-gallery-quadruped}
\end{figure}

\begin{figure}[H]
    \centering
  \includegraphics{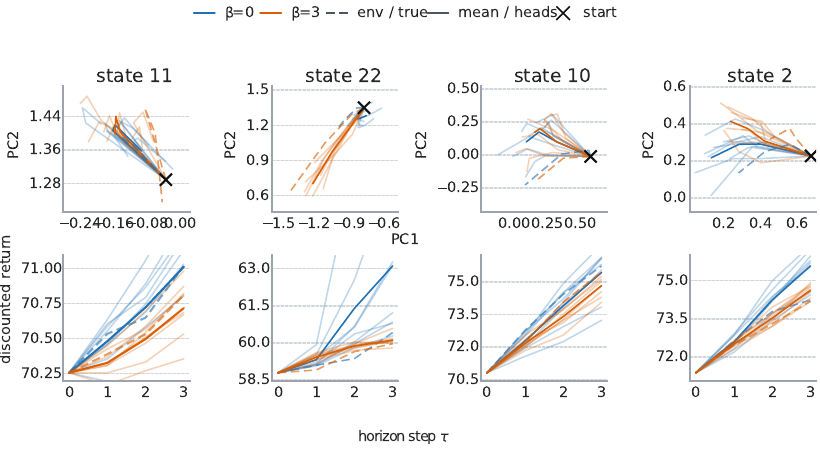}
  \caption{\textbf{Latent trajectory gallery on dog-stand at 50k (four states).} Same as Figure~\ref{fig:app-latent-gallery-reacher} but for dog-stand. }
    \label{fig:app-latent-gallery-dog}
\end{figure}

Across all three tasks, the state-level gallery makes the uncertainty effect concrete. Even when the predicted return stays competitive, the pessimistic planner tends to keep the per-head rollout bundle tighter than the optimistic planner.

\subsection{Task visualization}
\label{sec:Task_Visualization}

\begin{figure}[H]
\centering
\includegraphics{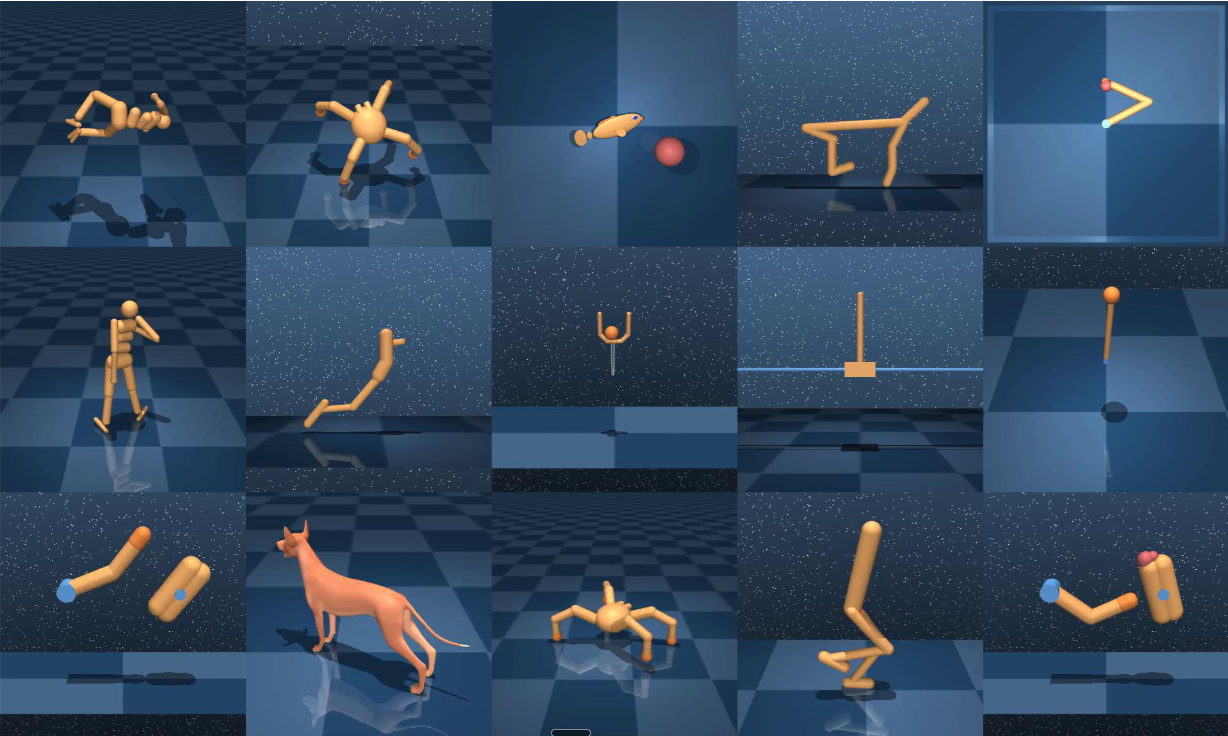}
\caption{\textbf{DMControl tasks visualization.} Images of all the embodiments we control in the DMControl tasks. The tasks include controlling them to run, walk, jump, balance, reach, and perform actions like swing-up and spin, covering a diverse range of continuous control scenarios.}
\label{fig:dmc_visual}
\end{figure}

\begin{figure}[H]
\centering
\includegraphics{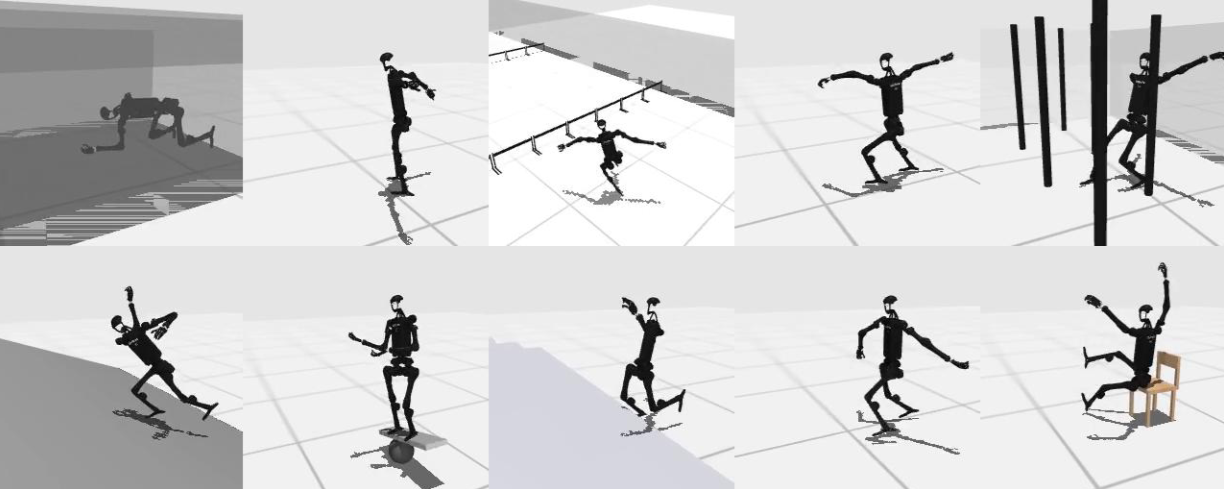}
\caption{\textbf{HumanoidBench visualization.} Images of the Unitree robot we control on HumanoidBench. The tasks include running, walking, crawling, balancing, sitting, reaching, and performing actions like walking on stairs or walking while avoiding collisions with poles.}
\label{fig:humanbench_visual}
\end{figure}

\section{Training losses}
\label{app:training-losses}

One update has two backward passes. The first trains the encoder, the dynamics heads, the reward head, and the value heads from one weighted loss. The second trains only the policy, on latents detached from that graph. The target value network $\bar{V}$ is an exponential moving average of the online value heads and receives no gradient.

A replay window is a short trajectory $(s_{t+u}, a_{t+u}, r_{t+u})$ together with the following observation and the stored planner distribution at each state. The encoder maps each observation to $z_{t+u} = h_{\theta}(s_{t+u})$. Dynamics heads have separate parameters. The reward head and the value heads are shared. Each dynamics head $i$ rolls out the replay actions from the encoded start state,
\begin{equation}
\hat{z}^{i}_{t} = z_t,
\qquad
\hat{z}^{i}_{t+u+1} = f_{\theta,i}(\hat{z}^{i}_{t+u}, a_{t+u}).
\end{equation}
Reward and value losses are applied to every latent of this rollout, including $\hat{z}^{i}_{t} = z_t$. Later true latents are stop-gradient targets. Each loss below is averaged over the minibatch and reduced over the window by $\sum_u \rho^{u} L_u$.

\textbf{Consistency.}
Each head is trained by mean squared error onto the encoded next latent,
\begin{equation}
\mathcal{L}_{\mathrm{cons}}
=
\sum_{u} \rho^{u}
\frac{1}{N_f}\sum_{i=1}^{N_f}
\big\|
\hat{z}^{i}_{t+u+1} - \mathrm{sg}(z_{t+u+1})
\big\|_2^2 .
\end{equation}
This updates $f_{\theta,i}$ along its own rollout and updates $h_{\theta}$ through $z_t$.

\textbf{Reward.}
The shared reward head predicts the environment reward from each head's latent by cross-entropy,
\begin{equation}
\mathcal{L}_{\mathrm{rew}}
=
\sum_{u} \rho^{u}
\frac{1}{N_f}\sum_{i=1}^{N_f}
\mathcal{L}_{\mathrm{CE}}\!\big(
R_{\theta}(\hat{z}^{i}_{t+u}, a_{t+u}),\, r_{t+u}
\big).
\end{equation}
This updates $R_{\theta}$, and the gradient flows back through $\hat{z}^{i}$ into the dynamics heads and the encoder. The reward target is the environment reward.

\textbf{Value.}
The value target of a true latent $z$ is a one-step policy backup, and it is constant during the model backward. Sample $a \sim \pi_{\theta}(\cdot \mid z)$ and detach the action. Average the target value heads, then average the backup over dynamics heads:
\begin{equation}
y(z)
=
\mathrm{sg}\!\left(
R_{\theta}(z, a)
+
\gamma\,
\frac{1}{N_f}\sum_{i=1}^{N_f}
\bar{V}_{\theta}\!\big(f_{\theta,i}(z, a)\big)
\right).
\end{equation}
The same target $y(z_{t+u})$ is used for every dynamics head. Every value head is trained on every head's rollout latent to match it,
\begin{equation}
\mathcal{L}_{\mathrm{val}}
=
\sum_{u} \rho^{u}
\frac{1}{N_v N_f}
\sum_{j=1}^{N_v}
\sum_{i=1}^{N_f}
\mathcal{L}_{\mathrm{CE}}\!\big(
V_{\theta,j}(\hat{z}^{i}_{t+u}),\, y(z_{t+u})
\big).
\end{equation}
The cross-entropy is computed per dynamics head and then averaged, so each head is trained to be value-accurate on its own latent. This updates $V_{\theta,j}$, and the gradient flows back through $\hat{z}^{i}$ into the dynamics heads and the encoder. Because $y(z)$ is stopped, the target network and the reward and dynamics evaluations inside the backup receive no gradient from $\mathcal{L}_{\mathrm{val}}$.

The model loss is a weighted sum of $\mathcal{L}_{\mathrm{cons}}$, $\mathcal{L}_{\mathrm{rew}}$, and $\mathcal{L}_{\mathrm{val}}$. One backward applies the gradient flow above.

\textbf{Policy.}
The rollout latents are then detached, and the policy is distilled toward the stored planner distribution by
\begin{equation}
\mathcal{L}_{\pi}
=
\sum_{u} \rho^{u}
\Big(
D_{\mathrm{KL}}\!\big(
\pi_{\theta}(\cdot \mid \mathrm{sg}(\hat{z}_{t+u}))
\,\big\|\,
\pi_{\mathrm{plan}}
\big)
-
\alpha\,
\mathcal{H}(\pi_{\theta})
\Big),
\end{equation}
averaged over dynamics heads. A second backward updates only $\pi_{\theta}$. The target value parameters are then moved toward the online value parameters by an exponential moving average.

\end{document}